\documentclass[10pt,twocolumn,letterpaper]{article}

\usepackage[pagenumbers]{cvpr} 

\definecolor{cvprblue}{rgb}{0.21,0.49,0.74}
\usepackage[pagebackref,breaklinks,colorlinks,citecolor=cvprblue]{hyperref}

\title{InverseMeetInsert: Robust Real Image Editing via Geometric Accumulation Inversion in Guided Diffusion Models} 

\author{Yan Zheng\\
UT Austin\\
{\tt\small yanzheng@utexas.edu}
\and
Lemeng Wu\\
UT Austin\\
{\tt\small lmwu@cs.utexas.edu}
}

\begin{document}
\maketitle
\begin{abstract}
In this paper, we introduce Geometry-Inverse-Meet-Pixel-Insert, short for GEO, an exceptionally versatile image editing technique designed to cater to customized user requirements at both local and global scales. Our approach seamlessly integrates text prompts and image prompts to yield diverse and precise editing outcomes. Notably, our method operates without the need for training and is driven by two key contributions: (i) a novel geometric accumulation loss that enhances DDIM inversion to faithfully preserve pixel space geometry and layout, and (ii) an innovative boosted image prompt technique that combines pixel-level editing for text-only inversion with latent space geometry guidance for standard classifier-free reversion. Leveraging the publicly available Stable Diffusion model, our approach undergoes extensive evaluation across various image types and challenging prompt editing scenarios, consistently delivering high-fidelity editing results for real images.

\end{abstract}    
\section{Introduction}
\label{sec:intro}

In recent years, advancements in image editing techniques, particularly those utilizing text-guided diffusion models, have made significant strides in real-world image editing. Our work introduces a new image editing method and framework that offers exceptional control and flexibility.

The diffusion model has been instrumental in advancing the generation of complex and detailed images. This model stands out for its ability to incorporate various types of information at each step of the image denoising process. One noteworthy implementation of this model is the text-to-image diffusion model, which excels at creating images that closely match natural language inputs, often referred to as 'text prompts.' Control techniques, developed based on these models, play a crucial role in enhancing image quality and relevance to better meet user preferences. These techniques often involve modifying textual prompts to refine the direction of image generation or using tools like bounding boxes, masks, and sketches for precise spatial guidance.

However, the application of these control techniques to real-world image editing is a relatively unexplored territory. One challenge arises from the inherent nature of diffusion models, which generate images from noisy Gaussian inputs. Identifying the corresponding noisy latent space that can accurately reconstruct real images has proven to be a formidable task. Although Denoising Diffusion Implicit Model (DDIM) helps convert real images into a suitable noisy latent space for text-to-image conditional diffusion models, image quality often suffers due to cumulative errors introduced by the basic ODE solver. This limitation hampers the direct application of established diffusion control techniques to practical image editing tasks.

Another issue in inversion-based techniques is stability. While classifier-free sampling is known to produce higher-quality images, incorporating null text weights into DDIM inversion can lead to instability. Null-text inversion and similar methods optimize text embedding rather than the text conditioning component to achieve better reconstruction. However, this optimization is computationally expensive for each new image, and the reconstruction of unedited parts remains unstable after inserting complicated text prompts.

To address these challenges, we introduce a novel concept called the 'geometric accumulative loss' for inversion. This loss leverages predicted image direction from the starting point back to an intermediate step. Instead of relying solely on text-based noise prediction, our geometric accumulative loss incorporates classifier-free guidance for the initial approximation. This approach capitalizes on the insight that achieving a precise and stable inverse path under classifier-free guidance for any real image would enhance editing stability.

The geometric accumulative loss finds a balance between these considerations by considering the input-encoded image latent as a reference point for fitting predictions, aiding the inversion process in retaining the geometric features of the input image. Our method allows for preliminary pixel-level editing, whether manual or automatic, followed by a reverse process equipped with the geometric accumulative loss. This process effectively translates coarse geometric information back to an intermediate step before resuming the reversal with standard classifier-free guidance. Our findings indicate that the geometric accumulative loss better preserves details in unedited areas compared to conventional DDIM inversion, as it is tailored to fit predictions under classifier guidance rather than text-only conditions.

To summarize, our method presents several novel contributions and advantages, including:
\begin{itemize}
    \item Our method allows users to perform precise and multi-area editing by inputting text prompts of any length and describing objects. This approach effectively eliminates the issue of word contamination commonly associated with the CLIP model.
    \item Our method effectively preserves background details in areas not being edited through a novel loss term, named as the geometrically accumulative loss for inversion that is specifically designed for simplicity and ease of implementation. This fast, plug-and-play loss term does not require tuning the model weights, thus it avoids compromising the pre-trained model's integrity and negates the need for replicating the model for each individual image.
    \item Our approach efficiently creates multiple edited images that accurately reflect the guidance from user-specified text prompts. It also enables more precise adjustments in visual details like color and geometric outline, further enhanced by our unique geometric accumulative loss.
\end{itemize}

\begin{figure*}[!bhpt]
  \centering
  \includegraphics[width=\textwidth]{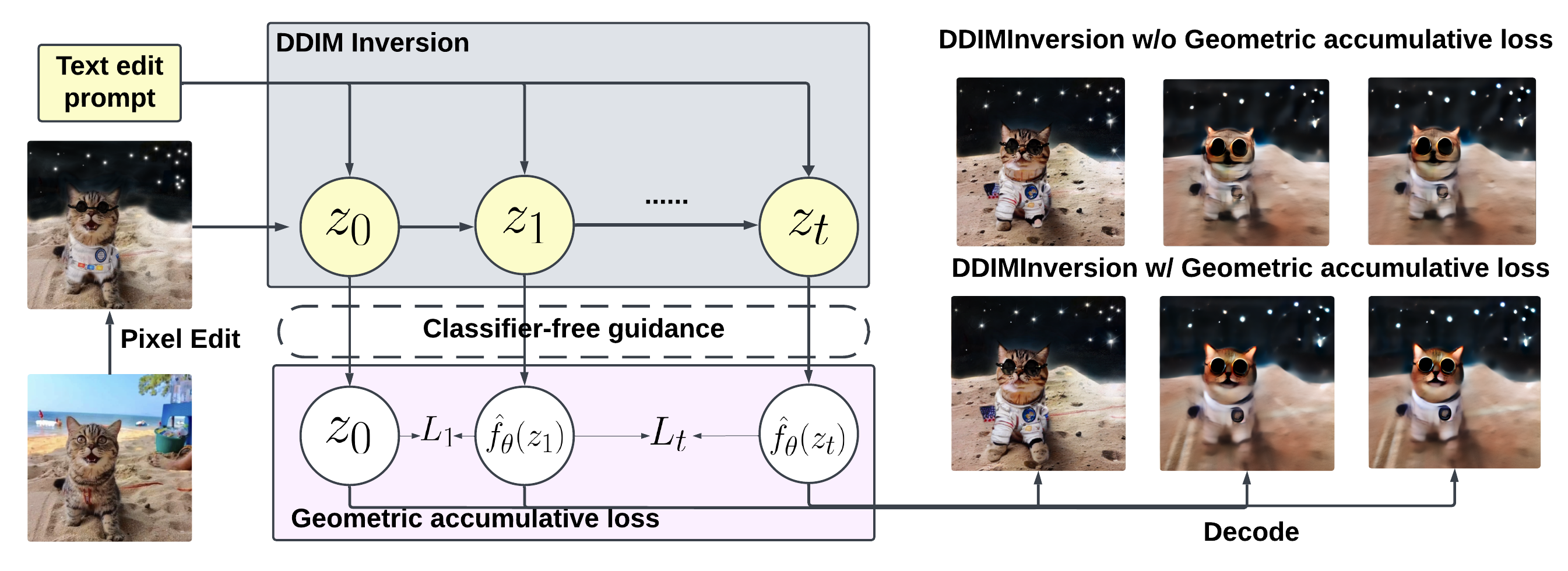}

  \caption{Pipeline of GEO. We first take a pixel level causal edit by user and a text edit prompt as input. The DDIM inversion process revert the pixel edited image back to latent, during this process, we apply Geometric accumulative loss to retain the latent information from both the pixel edit space and the text prompt guidance, compared with DDIM inversion, we get a better decoding image during the inversion process and as the result we get a fine-detailed result compared with Naive DDIM inversion.}
  \label{fig:pipeline}
\end{figure*}

\section{Related Work}
\paragraph{Text Condition Image Generative model}

The field of condition generative models has undergone a remarkable transformation, marked by a shift from early GAN-based and VAE-based models to the more sophisticated diffusion-based models. Initially, GAN based methods~\cite{crowson2022vqgan, liu2021fusedream,zhou2021lafite} set the foundation, offering impressive diversity and quality in image generation, yet they often fell short in accurately translating complex textual descriptions into images. This gap was significantly bridged by diffusion-based models~\cite{sohl2015deep, ho2020denoising, songscore, song2021maximum, song2020improved, song2019generative, karraselucidating, dhariwal2021diffusion, hoogeboom2023simple}. These newer models excel in synthesizing photorealistic images that closely describe user prompts, thanks to advancements in deep learning and their training on extensive data sets. Beyond their enhanced fidelity to text prompts, these models have also expanded the horizons of generative applications, finding use in diverse fields such as 3D modeling, novel view synthesis, and even music generation. This evolution not only represents a leap in the technical capabilities of text-to-image synthesis but also reshapes the potential of artificial intelligence in creative domains, providing tools that can transform text into vivid, accurate visual representations with unprecedented ease and flexibility, thereby setting new standards and opening new possibilities in the realm of digital art and beyond.

\paragraph{Diffusion Model for Image Editing} 
In recent years, the development of diffusion models  has introduced a more flexible design space for image editing tasks compared to Generative Adversarial Networks (GANs). These diffusion models offer a simpler training setup, exemplified by methods like SDEdit \cite{meng2021sdedit} and ILVR \cite{choi2021ilvr}.

Some approaches, exemplified by Textual Inversion \cite{textualinversion} and Dream-Booth \cite{ruiz2022dreambooth}, have showcased their proficiency in generating diverse images characterized by unique object attributes. They achieve this by fine-tuning diffusion models using multiple images, effectively identifying the inverse semantic latent representations that capture the distinct characteristics of objects in the embedding space. Similarly, Imagic\cite{kawar2023imagic} and UniTune\cite{valevski2022unitune}, leveraging the powerful Imagen model [43], have showcased impressive editing performance. However, a common limitation of these methods lies in their requirement for restrictive fine-tuning of pre-trained models, which can hinder their ability to fully harness the generalization potential of the models, often leading to issues such as overfitting or language drift.

Another category of methods, such as those represented by \cite{avrahami2022blended, nichol2021glide}, rely on user-provided masks to guide the diffusion process. While effective, this requirement for user-provided masks can limit the interactivity of these methods.

In the pursuit of text-only interactive editing, recent developments have given rise to optimization-free methods. For instance, Prompt-to-Prompt\cite{p2p} and DiffEdit\cite{couairon2022diffedit} have been proposed to automatically infer masks before initiating the editing process.

Prompt-to-Prompt (PTP)\cite{p2p} achieved comprehensive text-guided image editing without the need for diffusion model refinement. This method encompassed local editing even without a predefined mask. However, PTP primarily focused on generated image editing, asserting that the straightforward step-by-step inversion process is less reliable for real images, particularly when employing larger classifier-free guidance scales.
Null-text inversion (NTI) \cite{nulltext} introduced a strategy involving the modification of constant null-text embeddings into image-specific optimal embeddings. This approach was designed to achieve precise image reconstruction and subsequently applied Prompt-to-Prompt (PTP) techniques for real image editing. Its successor, Negative Prompt Inversion \cite{negtext}, attempted to expedite the null-text tuning phase of NTI for faster editing. However, both approaches essentially address the same optimization target, inheriting the constraints associated with null-text inversion. These constraints demand careful word selection in editing prompts, potentially limiting robustness.

In our method, we adopt an inversion-based technique for image editing while incorporating an image prompt derived directly from user-level pixel editing. We demonstrate that the combination of image prompts and text prompts for guidance during the inversion process can yield more robust and diverse editing results. Furthermore, we propose a novel, straightforward, and tailored loss function for inversion to ensure high-quality reconstructions.

\section{Method}

Our objective is to take a real-world image, referred to as $I$, and edit it to meet user specifications, yielding a modified image $I^*$. Additionally, as the editing requirements are not rigid and should allow for creative generativity, our method aims to produce a set of images that adhere to these editing criteria, collectively labeled as $\{ I^*_i \mid i = 1, 2, \ldots, n \}$.

\subsection{Background and Preliminaries}

\paragraph{Text-guided Latent Diffusion Models.} Consider a text-guided diffusion model that initiates with a textual embedding $C$ and a random Gaussian noisy image, denoted as $I_T$. Note that $C=\mathcal{T}(P)$ is the embedding of text prompt $P$ in natural language projected by the text encoder $\mathcal{T}$ and $I_T \in \mathbb{R} ^{H \times W \times C}$ is characterized by Gaussian i.i.d pixel values. 

The goal of the diffusion model is to progressively denoises the image, resulting in a sequence $I_T, I_{T-1},  \ldots, I_0$ such that $I_0$ corresponds to the text prompt $P$. But to allow the diffusion model to operate on a more compact representation, which reduces both time complexity and memory usage, Latent Diffusion Models (LDMs) ~\cite{ldm} uses an encoder $\mathcal{E}$ to map a given image $I$ into a latent embedding $z$. Subsequently, a decoder $\mathcal{D}$, is employed to reconstruct the input image from $z$, such that $\mathcal{D}(\mathcal{E}(I)) \approx I$. Therefore, we only need to replace the image $I$ with its latent embedding $z$ in the follwing algorithm.

$z_t$ is a sample with added standard Gaussian noise $\epsilon$, where the noise is introduced according to a time-dependent schedule $\alpha_t$. The relation is given by $z_t = \sqrt{\alpha_t} z_0 + \sqrt{1 - \alpha_t} \epsilon$, corresponding to a forward diffusion process:
\begin{equation}
    q(x_t | x_{t-1}) := \mathcal{N}(x_t; \sqrt{\alpha_t}x_{t-1}, (1 - \alpha_t)I)
\end{equation}

To learn the reverse process, the mean value $\mu_{\theta}(x_t, t)$, parameterized by $\theta$, is predicted:
\begin{equation}
    p_{\theta}(x_{t-1} | x_t) := \mathcal{N}(x_{t-1}; \mu_{\theta}(x_t, t), \Sigma_{\theta})
    \label{eq:reverse}
\end{equation}

A denoiser module, parameterized by the network $\epsilon^{\theta}$, is utilized for fitting the objective, which is equivalent to fitting the mean value prediction in the reverse process:
\begin{equation}
    \min_{\theta} \mathbb{E}_{z_0, \varepsilon \sim \mathcal{N}(0, I), t \sim \text{Uniform}(1, T)} \| \varepsilon - \varepsilon_\theta(z_t, t, C) \|^2. 
\end{equation}

During the inference phase, the model can employ either stochastic sampling as in DDPM~\cite{ddpm}, which introduces noise at each sampling step, or deterministic sampling as in DDIM~\cite{ddim}, which follows an ODE-like deterministic trajectory. Specifically, we utilize the deterministic DDIM sampling approach to leverage the inverse characteristics of the ODE path:
\begin{equation}
    \frac{z_{t-1}}{\sqrt{\alpha_{t-1}}} = \frac{z_{t}}{\sqrt{\alpha_{t}}} + (\frac{\sqrt{1-\alpha_{t-1}}}{\sqrt{\alpha_{t-1}}} - \frac{\sqrt{1-\alpha_{t}}}{\sqrt{\alpha_{t}}}) \epsilon^{\theta}(z_t ,t, C)
\end{equation}

\paragraph{DDIM inversion.} The challenge in diffusion-based inversion lies in transforming a real image into its Gaussian noisy counterpart by reversing the diffusion process. Essentially, this process operates inversely, transitioning from $z_0$ to $z_T$ rather than the typical $z_T$ to $z_0$ pathway. Here, $z_0$ represents the encoded form of the provided real image $I_0$. A straightforward inversion method for DDIM sampling was proposed in references [13, 35], relying on the idea that the ODE process can be inverted using very small steps:
\begin{equation}
    \frac{z_{t+1}}{\sqrt{\alpha_{t+1}}} = \frac{z_{t}}{\sqrt{\alpha_{t}}} + (\frac{\sqrt{1-\alpha_{t+1}}}{\sqrt{\alpha_{t+1}}} - \frac{\sqrt{1-\alpha_{t}}}{\sqrt{\alpha_{t}}}) \epsilon^{\theta}(z_t ,t, C)
\end{equation}

\paragraph{Classifier-free Guidance.}
Classifier-free guidance~\cite{classifierfree} addresses the challenge of amplification of the effect induced by the conditioned text in text-guided generation. This technique involves making predictions both conditionally and unconditionally. These predictions are then combined. Formally, let $\emptyset = \mathcal{T}("")$ represent the embedding of an empty text and let $\omega$ denote the guidance scale parameter. The classifier-free guidance prediction is then defined as
\begin{equation}
    \tilde{\varepsilon}_\theta(z_t, t, C, \emptyset) = w \cdot \varepsilon_\theta(z_t, t, C) + (1 - w) \cdot \varepsilon_\theta(z_t, t, \emptyset)
\end{equation}
For example, in Stable Diffusion, the default value for $\omega$ is 7.5.


\subsection{Insert Meet Inverse}
The task of editing images in pixel space has a well-established history with conventional methods, offering greater flexibility. This is due to the ability to employ pixel-level operations such as manual drawing, pasting, layer masking, merging, and smoothing.
To this end, we propose \textbf{GEO}, in which we suggest that pixel-level editing can be efficiently integrated into the noisy latent space using a diffusion model-based inversion technique. This integration is then reversed back into the real image pixel space, resulting in more natural and semantically enhanced pixel-level edits.
\begin{figure*}[!bhpt]
  \centering
  \includegraphics[width=1.0\textwidth]{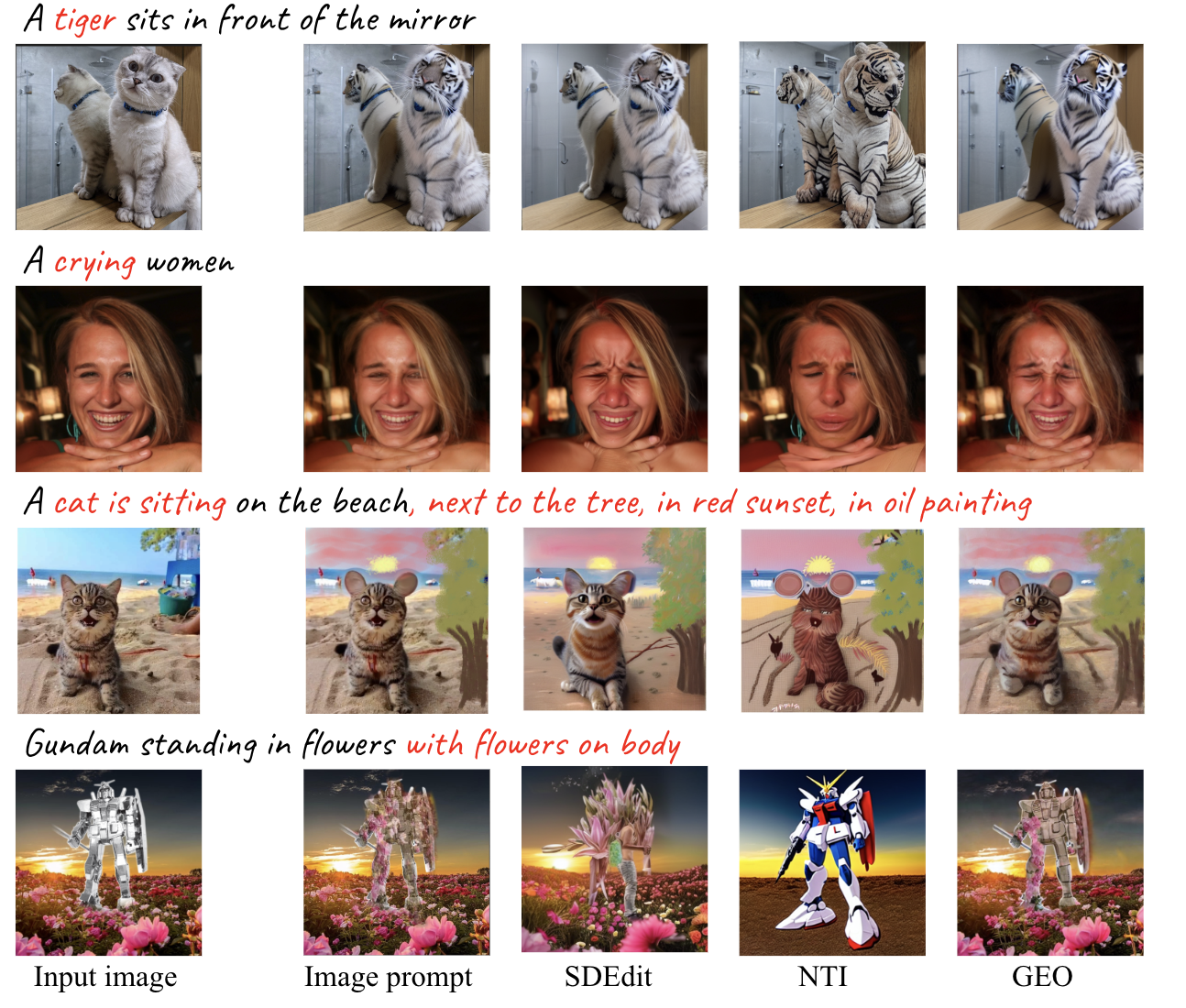}
  \caption{Editing examples of various of input images with different styles of Image prompt. We take image prompt from SDEdit, stickers, user stroke and brush. Combine with the text prompt input, GEO can refine the image prompt into high-fidelity image result that keeps the requirement from both image and text input.  } 
  \label{fig:gallery}
\end{figure*}
\subsubsection{Insert in Pixel Space: Editing Proposal in Pixel Space} 

We suggest various editing methods capable of creating initial editing proposals in pixel space that match the user-provided text prompts. Our method uniquely avoids modifying the text encoder~\cite{ptinversion} and cross-attention components in U-net~\cite{p2p, nulltext, negtext}. Consequently, there are no restrictions on the length or content of the text prompts, unlike methods based on attention mixture.

In the context of image editing, our approach begins with an unaltered real-world image, denoted as $I$. The user is then asked to provide two distinct text prompts: $P$, which aims to describe the original image $I$ as accurately as possible, and $P^*$, which outlines the desired modifications. Subsequently, the user can engage in pixel-level editing, employing a variety of elemental operations:

\noindent\textbf{Brush Stroke:} This function enables users to select specific regions within the image for editing. It is particularly useful for tasks such as altering colors or swapping backgrounds. The brush stroke tool offers a straightforward and intuitive interface for these modifications, allowing for precise control over the editing area.

\noindent\textbf{Image Paste:} Users can also incorporate elements from external sources into the image $I$. This involves selecting an object from a separate image and seamlessly integrating it into the original picture. The merging process can either be executed manually, involving steps like image mask cropping and layer overlapping, or automated through advanced mask segmentation tools, such as Segment Anything~\cite{seganything}. This feature is instrumental in enhancing the creative flexibility of the editing process, enabling users to blend elements from diverse sources effectively.

\noindent\textbf{SDEdit:} The Stochastic Differential Equation (SDE) editing method is particularly adept at generating intricate object details or styles that are challenging to manually draw or derive from other images. Initially, the input image is converted into a latent space representation through an encoder, denoted as $z =\mathcal{E}(I)$. This latent representation, $z$, is then subjected to the addition of independent standard Gaussian noise, expressed as $z_t = \sqrt{\alpha_t} z + \sqrt{1 - \alpha_t} \epsilon$. The denoising process commences not from the initial step $T$ but from an intermediate step $t$. This approach constitutes a stochastic inversion process that does not reliably reconstruct $z$ from $z_t$ when $t$ is close to $T$. Consequently, the reversed image, while not directly suitable as an output for editing results, can be superimposed onto the original image $I$ with appropriate opacity to provide visual guidance from the pixel space.

It is important to note that these three operations – Brush Stroke, Image Paste, and SDE Editing – can be combined in various ways to achieve more refined and tailored editing results. This flexibility allows for an extensive range of creative possibilities, offering users the ability to fine-tune their edits to align closely with their vision and the guidance provided by the text prompts.

\subsubsection{Inverse in Latent Space: Geometric Accumulation Inversion}

A distinguishing feature of the denoising process is evident from Eq.(\ref{eq:reverse}), which enables a direct estimation of $z_0$ from the model's denoising direction $\epsilon_{\theta}$:
\begin{equation}
    \hat{z}_0 = \frac{1}{\sqrt{{\alpha}_t}} \left( z_t - \sqrt{1 - {\alpha}_t}\varepsilon_\theta(z_t, t) \right) 
\end{equation}

During the entire generation process from timestep $T$ to $0$, we find that the geometric layout and outlines are predominantly determined in the initial steps near $T$. As a result, the predicted latent representation undergoes significant changes early on, both in the latent and decoded image spaces. In subsequent steps, once the geometric structure is relatively established, the focus shifts to refining details in blurred regions to enhance the image's realism.

Some studies have attempted to utilize this early predicted image information to formulate mask-based comparison losses for tasks like inpainting or image restoration. Nonetheless, the tendency for early predictions to appear blurred or overly smooth renders them unsuitable for direct comparison with real input images. Such comparisons can lead to overly smooth and unrealistic alterations, deviating significantly from the user's original image. Thus, there is a need for an approach that harnesses the directional information provided by these predictions within the inversion process, which is essential for the editing workflow, rather than solely considering the denoising direction during inference.

Another issue in inversion-based techniques is stability, even though classifier-free sampling is known to yield better quality images. When incorporating null text weights into the DDIM inversion, the process can become unstable. Null-text inversion and subsequent methods optimize text embedding rather than the text conditioning component to achieve better reconstruction from timestep $T$. However, this optimization is computationally expensive for each new image, and the reconstruction of the unedited parts remains unstable post-editing with the inserted text prompt.

To address this, we introduce a geometric accumulative loss for inversion, capitalizing on the predicted image direction from timestep $0$ back to an intermediate step $t$. In each inverse step, starting with $z_t$, we first apply a text-only DDIM inversion to obtain a backward direction $v^{0}_{t} = \epsilon^{\theta}(z_t ,t, \mathcal{T}(P^{*}))$ as an initial estimate. We then refine this initial point by fitting the objective defined as the geometric accumulative loss:

\begin{equation}
    \min_{v_t} \| \tilde{f}^{\theta}( z_{t+1}(v_t) ) - \tilde{f}^{\theta}(z_t) \|
    \label{eq:geoaccum}
\end{equation}
where $z_{t+1}(v_t) =  \sqrt{\alpha_{t+1}} \left(\frac{z_{t}}{\sqrt{\alpha_{t}}} + \left(\frac{\sqrt{1-\alpha_{t+1}}}{\sqrt{\alpha_{t+1}}} - \frac{\sqrt{1-\alpha_{t}}}{\sqrt{\alpha_{t}}}\right) v_t \right)$ and $\tilde{f}^{\theta}$ is the prediction given by 
\begin{equation}
    \tilde{f}^{\theta}(z) = \frac{1}{\sqrt{{\alpha}_t}} \left( z - \sqrt{1 - {\alpha}_t} \tilde{\varepsilon}_\theta(z_t, t, \mathcal{T}(P^{*}), \emptyset) \right). 
\end{equation}
$\tilde{f}^{\theta}(z_0)$ is directly set as the encoded input image $\mathcal{E}(I_0)$.

Instead of relying on the conventional text-only noise predictor for $v_t$ initialization as in standard DDIM inversion, our geometric accumulative loss incorporates the classifier-free guidance prediction for the $z_0$ approximation. This stems from the insight that finding a precise and stable inverse path under classifier-free guidance for any real image would enable standard Prompt-to-Prompt editing stability. However, achieving this precision is challenging through straightforward optimization or model tuning. While text-only inversion is simple and can be improved iteratively for better reconstruction, it does not address the issue of classifier-free guidance altering the input image in the reverse process, which could also degrade the image quality due to amplified text condition effects.

The geometric accumulative loss represents a balance between these considerations. It leverages the fact that for image generation at timesteps close to $0$, the predicted $\hat{z}_0$ closely approximates the final output. Inversely, we can consider the input encoded image latent $z_0 = \mathcal{E}(I_0)$ as a reference point for fitting predictions, aiding the inversion process in retaining the geometric features of the input image. However, it is impractical for the predicted latent at each inversion step to align with $z_0 = \mathcal{E}(I_0)$ due to varying noise levels across timesteps, particularly from timestep $t$ to $0$. But given the denoiser network's inherent smoothness, predictions should not significantly differ from timestep $t$ to $t-1$. Therefore, a practical method to transfer prediction information from timestep $0$ is to compare it with the previous step, as indicated in Eq.(\ref{eq:geoaccum}).

A paradox of text-conditioned diffusion models is their tendency to amplify text condition information from the visual cues in noisy image space. We exploit this property to first propose a preliminary editing in pixel space, whether manually or automatically through masking and pasting. Then, the reverse process, equipped with our geometric accumulative loss, effectively translates this coarse geometric information back to an intermediate step. Subsequently, we resume the reversal with standard classifier-free guidance. We find that our geometric accumulative loss better preserves details in the unedited area compared to plain DDIM inversion, as it is tailored to fit predictions under classifier guidance rather than text-only conditions.

\begin{algorithm}
\caption{Geometrically accumulative inversion}
\begin{algorithmic}[1]
\State \textbf{Input:} An edit prompt embedding $C = \mathcal{T}(P^{*})$, a stop time $t$, and input image $\hat{I}$ with rough pixel editing.
\State \textbf{Output:} Noise vector $z_t$.

\For{$t = t, t-1, \ldots, 0$}
\State $v^{0}_{t} = \epsilon^{\theta}(z_t ,t, \mathcal{T}(P^{*}))$
    \For{$i = 0, \ldots, N-1$}

        \If{$t = 0$}
            \State $v^{i+1}_{0} \leftarrow v^{i}_{0} - \nabla_{v_{0}} \| \tilde{f}^{\theta}( z_{1}(v^i_0) ) - \mathcal{E}(\hat{I}) \|^2$
        \Else
            \State $v^{i+1}_{t} \leftarrow v^{i}_{t} - \nabla_{v_{t}} \| \tilde{f}^{\theta}( z_{t+1}(v^i_{t}) ) - \tilde{f}^{\theta}( z_{t} ) \|^2$
        \EndIf  
    \EndFor

    \State Set $z_{t+1} \leftarrow  \sqrt{\alpha_{t+1}} \left(\frac{z_{t}}{\sqrt{\alpha_{t}}} + \left(\frac{\sqrt{1-\alpha_{t+1}}}{\sqrt{\alpha_{t+1}}} - \frac{\sqrt{1-\alpha_{t}}}{\sqrt{\alpha_{t}}}\right) v_t^N \right)$
        
\EndFor

\State \textbf{return} $z_t$
\end{algorithmic}
\label{alg:geoinv}
\end{algorithm}

\begin{algorithm}
\caption{Geometric Accumulative Editing}
\begin{algorithmic}[1]
\State \textbf{Input:} An edit prompt embedding $C = \mathcal{T}(P^*)$ and an input image $I$.
\State \textbf{Output:} An edited image $I^*$ aligned with $P^*$.

\State Perform manual pixel-level editing using techniques such as brush strokes, image pasting, or selective edits. Denote the resulting image in pixel space as $I^*$.
\State Encode the image into a latent space: $z_0 \leftarrow \mathcal{E}(I^*)$.
\State Set a stopping time $t$ and apply Algorithm \ref{alg:geoinv} to perform the inversion from $z_0$ to $z_t$.
\State Initialize the guidance scale $\omega = 7.5$ to initiate classifier-free guidance on $z_t$ from timestep $t$ with the edit prompt embedding $C = \mathcal{T}(P^*)$: $z_t, z_{t-1}, \ldots, z_0$.
\State Decode back to the pixel space: $I^{*} \leftarrow \mathcal{D}(z_0)$.
\State Return the edited image $I^{*}$.
\end{algorithmic}
\label{alg:geoedit}
\end{algorithm}

The workflow of our approach is illustrated in Figure \ref{fig:pipeline}. The algorithms are presented in Algorithm \ref{alg:geoinv} and Algorithm \ref{alg:geoedit}. It is noteworthy that we commonly select a stop time of $t = 25$ as the optimal choice.

\section{Applications}

\subsection{Experiment Set Up}

In our experimental setup, we employ the text-conditional Latent Diffusion Model, also referred to as Stable Diffusion, which has been trained with 890 million parameters on the LAION-5B dataset at a resolution of 512 × 512. For the DDIM schedule, we adhere to a regimen consisting of 50 steps while retaining the original hyperparameter settings of Stable Diffusion. Notably, our inversion procedure can be completed within a duration of one minute and thirty seconds when executed on a single A100 GPU.

\subsection{Object Editing/ Style Editing}

As depicted in Figure \ref{fig:gallery}, our approach initiates the editing process by first modifying the input real image at the pixel level, utilizing it as an image prompt. Subsequently, we provide the user with guided, unlimited edit prompts in the form of natural language guidance for our inversion procedure. Our method excels at seamlessly swapping objects within an image, transitioning from one object, such as a 'cat,' to another, like a 'tiger.' Furthermore, our method demonstrates versatility in style editing, capable of adjustments at both local and global scales.

For instance, consider the 'A cat on the beach' image. After introducing the 'oil painting' guidance in the edit prompt, the style of the entire image is successfully transformed. This transformation smoothly merges the roughly brushed strokes with the remaining photorealistic portions. In the 'gundam' case, only the 'gundam' object's style is modified to harmonize with the image, leaving the background untouched.

It is noteworthy that our method not only accurately captures color information in pixel editing but also possesses the ability to discern correct semantic information from rough or ambiguous images. This capability is exemplified in the second example, where we transform an initially unrealistic 'smiling woman' image into a 'crying woman.' We achieve this by generating an initially unconvincing crying woman's face using SDEdit, which we then superimpose onto the original input image. We accomplish this by segmenting out the woman's face and adjusting the opacity of the overlaid face as an image prompt hint. Although the input image prompt may appear unrealistic to human observers, our method, when provided with the appropriate guidance, accurately steers the inversion process toward the correct semantic interpretation of `crying.'

While conducting local edits, our method excels in preserving the background due to its specific geometric accumulative design, which retains the geometric information of the original input image. In contrast, Null-text Inversion can sometimes maintain the background but lacks the fine control exhibited by our approach. Additionally, Null-text Inversion is sensitive to the prompt, as observed in the `gundam' case where the background undergoes a complete transformation. In other cases, such as `two cats' and `A cat on the beach,' the cat's appearance is semantically accurate but unsuitable and unreal. SDEdit generates images with more stable styles; however, its level of detail is less conducive to robust editing applications.

\subsection{Multiarea Editing}
\begin{figure}[!bpht]
  \centering

   \includegraphics[width=1.0\linewidth]{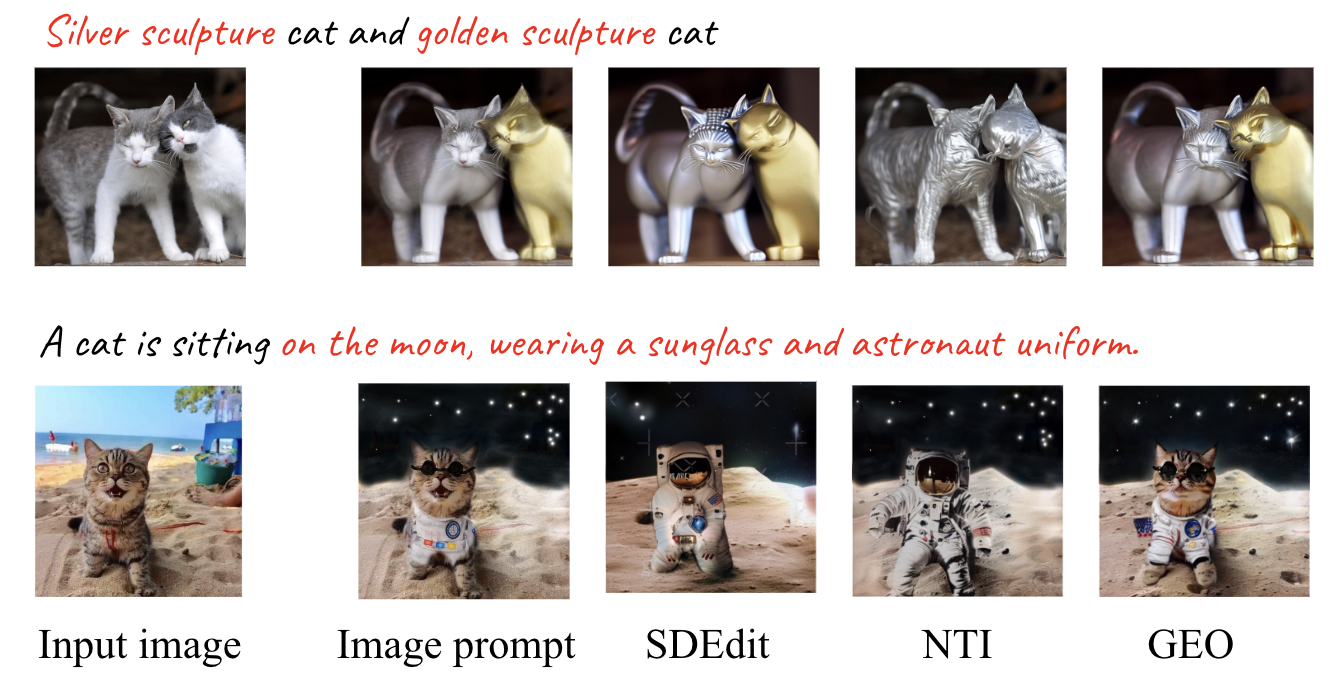}
    \caption{Multi-area editing result from different methods. Our method can accurately capture the multiarea edit requirement from image and text prompt while other methods tend to merge the different areas' concept together.}
   \label{fig:multi}
\end{figure}

As evident from the examples presented in Figure \ref{fig:multi}, our method incorporates an image prompt that offers flexible editing proposals derived from the image space, allowing for multi-area editing. For instance, this capability enables simultaneous editing of multiple regions within a single image, such as distinct materials for two cats and background, as well as the addition of glass and cloth elements. This addresses a long-standing issue prevalent in CLIP-based text-guided models, namely, the problem of contamination.

For illustration, consider the editing result of the 'silver sculpture cat and golden sculpture cat' example using null-text inversion. In this case, the presence of the word 'silver' inadvertently affects both cats, rather than producing two cats composed of separate materials. We contend that the self-amplification effect inherent in text-conditioned models can be harnessed to enhance the semantic information conveyed by the image prompt.

In other words, while the initial manual editing originating from the pixel space may fall short of achieving a realistic and natural appearance, during the inversion process, as we progress to an intermediate step and subsequently reverse our modifications, the less desirable aspects of the image prompt can be rectified under the guidance of the editing prompt. This correction process allows us to refine and align the image prompt more closely with the intended editing prompt.

\subsection{Customize Editing}

\begin{figure}[!bpht]
  \centering
  
   \includegraphics[width=1.0\linewidth]{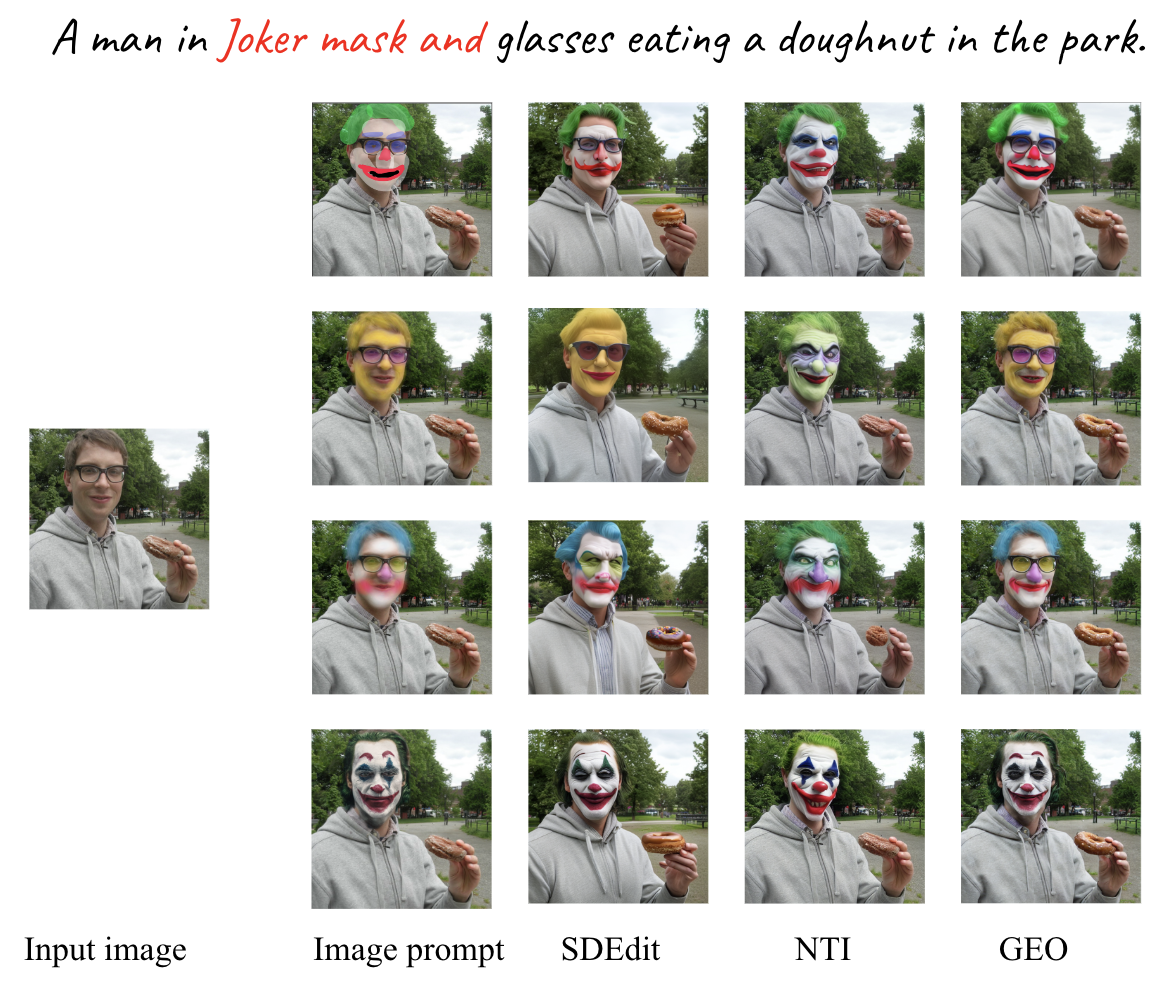}

   \caption{Custom editing result. For a same image and a same text prompt, we can easily change the style according to the input image prompt. This allows user to customize their own style with brushing, stroking, or even some sticker pasted from other image. }
   \label{fig:customize}
\end{figure}

Another advantageous aspect of incorporating the image prompt is its ability to offer users greater flexibility in customizing features such as precise color matching or seamlessly inserting objects from other images into the edited image. This flexibility is demonstrated in Figure \ref{fig:customize}, where our method excels at preserving the user-provided colors specified in the image prompt. Notably, it seamlessly integrates the Joker mask from a film poster onto a human face without distorting the original facial details. In contrast, both SDEdit and NTI introduce distortions to the facial features, whereas our results maintain the natural appearance of the original face.

\section{Ablations}

\begin{figure}[t]
  \centering
  
   \includegraphics[width=1.0\linewidth]{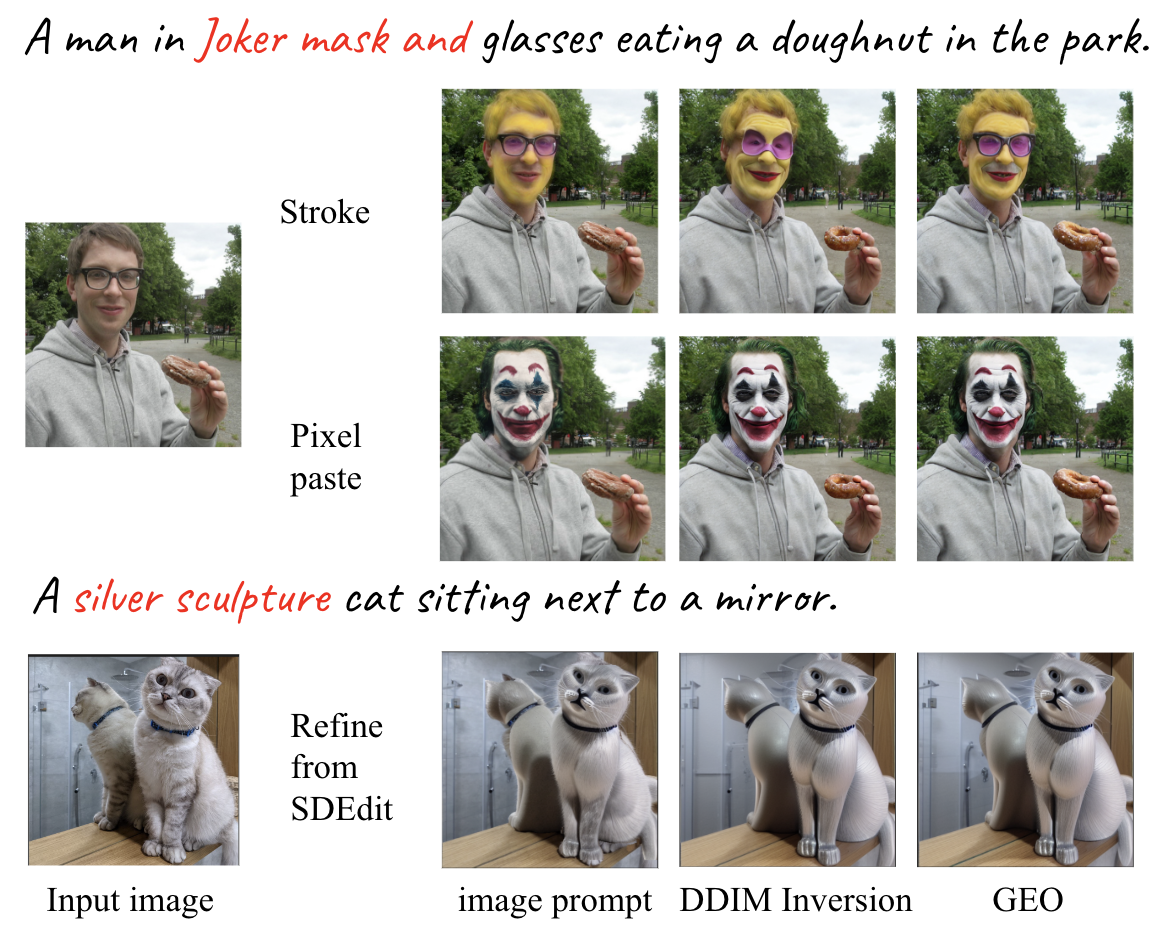}

   \caption{Ablation study on detail preserving ability between DDIO inversion and GEO. DDIM Inversion usually tends to distort the background during edtigin. Our GEO can keep edit the prompt area accurately while keeping the background area unchanged.}
   \label{fig:ablation}
\end{figure}

\begin{figure}[t]
  \centering
  
   \includegraphics[width=1.0\linewidth]{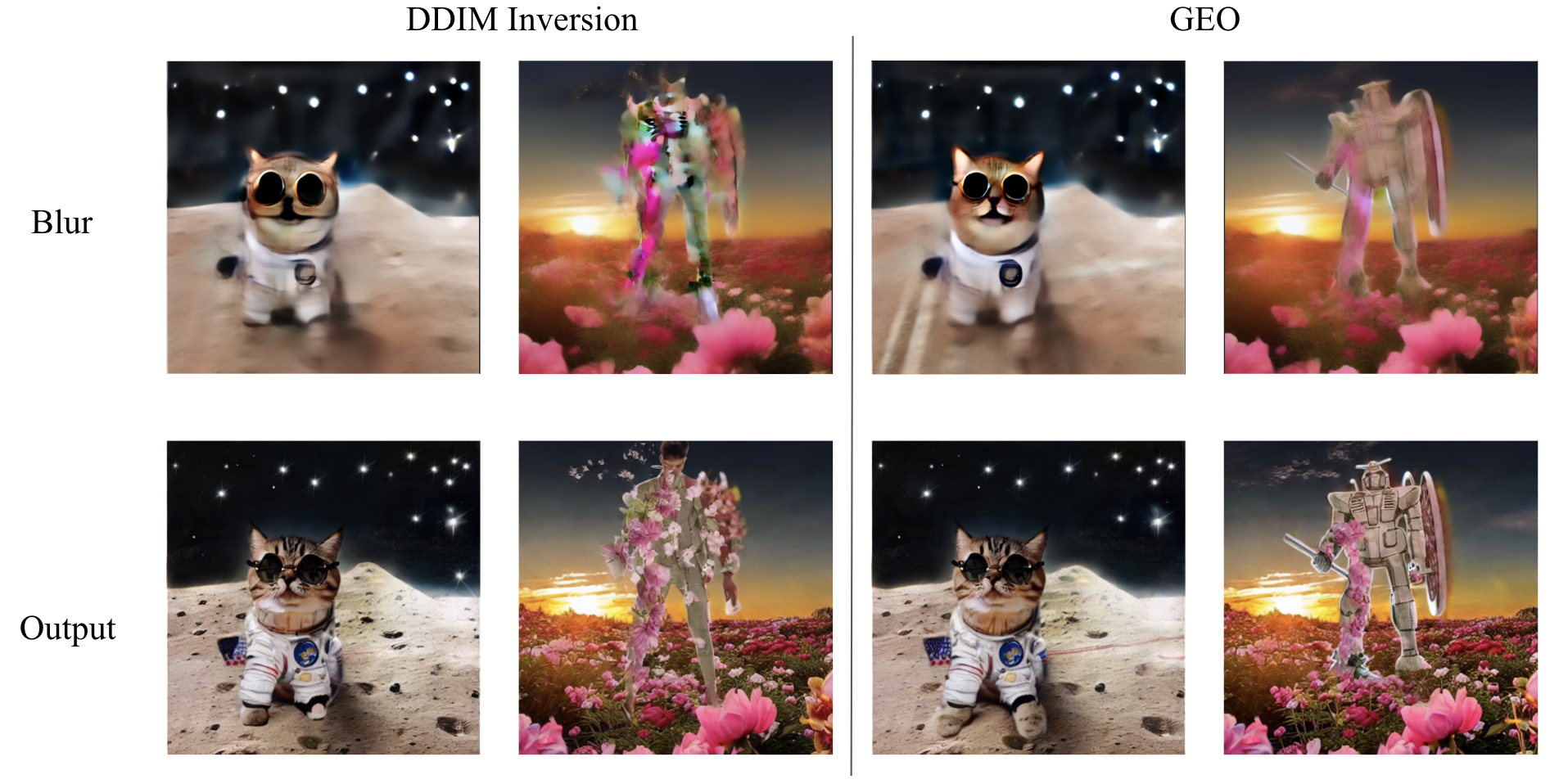}

   \caption{Visualization of the blur intermediate state. Compared of DDIM Inversion, our intermediate state keeps the layout of the image prompt, which can easily guided by text prompt for the final output result.}
   \label{fig:blur}
\end{figure}

Our method initializes the Geometric Accumulation Inversion using the DDIM backward step. To evaluate the effectiveness of our proposed loss, we conducted ablation experiments as shown in Figures \ref{fig:ablation} and \ref{fig:blur}.

In Figure \ref{fig:ablation}, we focus on the preservation of background details. It is evident that without our loss, the details of the hand and the donut in the background easily distort during the reverse generation process. However, when our loss is applied, the hand remains faithful to its original appearance after the reverse generation.

In Figure \ref{fig:blur}, we delve into the impact of the predicted latent space, visualized as a blurred image in the first row. It becomes clear that without proper preservation of shape information in the blurred image path, the resulting reconstructed image deviates from the desired outcome aligned with the guidance provided by the text prompt. Our method enhances the geometric memory within the blurred image path, resulting in a reconstructed image of higher quality that better corresponds to the editing prompt.


\section{Limitations}

Our current image prompt design relies on manual adjustments, which can introduce errors during pixel space masking. Additionally, the manual nature of image prompt creation limits the scalability of our method for batch processing, impacting quantitative metrics.

{
    \small
    \bibliographystyle{ieeenat_fullname}
    \bibliography{main}
}

\end{document}